\documentclass[10pt,twocolumn,letterpaper]{article}

\usepackage{cvpr}              % To produce the CAMERA-READY version
\definecolor{cvprblue}{rgb}{0.21,0.49,0.74}
\usepackage[pagebackref,breaklinks,colorlinks,allcolors=cvprblue]{hyperref}

\usepackage{multirow}

\def\paperID{} % *** Enter the Paper ID here
\def\confName{CVPR}
\def\confYear{2026}

\title{CATNet: Collaborative Alignment and Transformation Network \\ for Cooperative Perception}

\author{
Gong Chen$^{1}$ \quad
Chaokun Zhang$^{2}$\thanks{Corresponding author.} \quad
Tao Tang$^{1}$ \quad
Pengcheng Lv$^{3}$  \quad % 如果有多个机构
Feng Li$^{1}$  \quad
Xin Xie$^{1}$
\\
$^{1}$School of Computer Science and Technology, Tianjin University  \quad \\
$^{2}$School of Cybersecurity, Tianjin University  \quad
$^{3}$School of Future Technology, Tianjin University 
}

\begin{document}
\maketitle
\begin{abstract}
Cooperative perception significantly enhances scene understanding by integrating complementary information from diverse agents. However, existing research often overlooks critical challenges inherent in real-world multi-source data integration, specifically high temporal latency and multi-source noise. To address these practical limitations, we propose Collaborative Alignment and Transformation Network (CATNet), an adaptive compensation framework that resolves temporal latency and noise interference in multi-agent systems. Our key innovations can be summarized in three aspects. First, we introduce a Spatio-Temporal Recurrent Synchronization (STSync) that aligns asynchronous feature streams via adjacent-frame differential modeling, establishing a temporal-spatially unified representation space. Second, we design a Dual-Branch Wavelet Enhanced Denoiser (WTDen) that suppresses global noise and reconstructs localized feature distortions within aligned representations. Third, we construct an Adaptive Feature Selector (AdpSel) that dynamically focuses on critical perceptual features for robust fusion. Extensive experiments on multiple datasets demonstrate that CATNet consistently outperforms existing methods under complex traffic conditions, proving its superior robustness and adaptability.
\end{abstract}    
\section{Introduction}
\label{sec:intro}
Accurate environmental perception is paramount for the reliability of autonomous driving systems, as it directly influences downstream decision-making and control \cite{Survey1}. However, single-agent perception systems inherently face limitations due to restricted fields of view and prevalent occlusion phenomena. These constraints pose significant challenges \cite{gao2024survey, han2023collaborative, yazgan2024collaborative} in maintaining robust perception capabilities within large-scale scenarios. To overcome these constraints, multi-agent cooperative perception \cite{xu2022opv2v, xu2023v2v4real, hao2024rcooper} has emerged as a critical technological paradigm. By leveraging Vehicle-to-Everything (V2X) communication infrastructures to achieve distributed information fusion, it has become a prominent research focus in the field.

Although numerous studies \cite{xu2022v2xvit, xu2024v2x, xu2022opv2v} have demonstrated promising results through feature interaction modeling under ideal communication assumptions, their practical effectiveness in dynamic real-world environments \cite{gu2023feaco} remains constrained. Specifically, inter-agent communications are vulnerable to channel interference and transmission noise \cite{tang2025rocooper}, leading to feature distortion in vehicular perception data. Existing methods struggle to reconstruct spatio-temporally consistent perceptual representations \cite{yazgan2024survey}, resulting in substantial performance degradation. The fundamental challenges can be attributed to two critical issues:

% 摘要图
\begin{figure}[!t]
    \centering
    \includegraphics[width=0.45\textwidth]{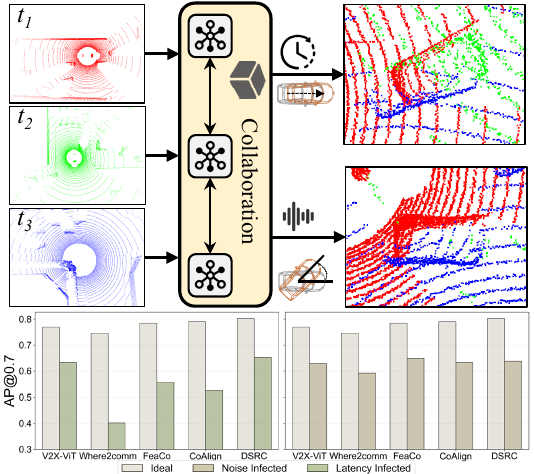}
    \caption{Schematic of cooperative perception under latency and noise.}
    \label{fig:abstract}
\end{figure}

\textbf{Time-Varying Communication Latency}. Dynamic latency poses a significant threat to system robustness by disrupting the spatio-temporal coherence between features from the ego-vehicle and its collaborators. As illustrated in the top-right panel of Figure \ref{fig:abstract}, features from different timestamps are misaligned due to transmission delays. This misalignment creates ghosting artifacts and feature fragmentation, fundamentally compromising perception consistency. The severity of this issue is quantified in our analysis (Figure \ref{fig:abstract}, bottom-left bar chart), where latency leads to a performance drop of up to 46\%. While prior solutions \cite{mrcnet, yang2023spatio} typically perform local temporal alignment on individual agent features before fusion, these frameworks prove insufficient as they inadequately leverage global spatio-temporal context.

\textbf{Noise-Induced Feature Degradation}. Features in cooperative perception are susceptible to corruption from multi-source noise during transmission. This degradation is exacerbated under asynchronous conditions, where coupled interference effects severely impact the pipeline. As depicted in the bottom-right panel of Figure \ref{fig:abstract}, noise deteriorates the geometric structure of the point cloud, introducing spurious artifacts and significant distortion of object shapes. This detrimental effect is confirmed in the bar chart, where noise pollution causes a performance drop of up to 17\%. Existing methods primarily focus on geometry-based registration \cite{coalign, huang2024roco, wang2020v2vnet} or fixed-threshold denoising \cite{hu2024adaptive, sheng2024semantic}. They either fail to adequately correct for systematic errors or inadvertently suppress crucial discriminative features, leading to suboptimal outcomes.

To address these challenges, we propose Collaborative Alignment and Transformation Network (\textbf{CATNet}), a dynamic adaptive compensation framework. It effectively tackles robust collaboration in multi-agent systems through the ingenious integration of three novel modules. 
% First, to counteract spatio-temporal misalignment from dynamic latency, we design the Spatio-Temporal Recurrent Synchronization (\textbf{STSync}) module. It establishes a global temporal context by recurrently propagating the ego-vehicle's features to iteratively align asynchronous data streams. 
To handle spatio-temporal misalignment, we design a Spatio-Temporal Recurrent Synchronization (\textbf{STSync}) module. It establishes a global temporal context by recurrently propagating the ego-vehicle's features to iteratively align asynchronous data streams. 
Subsequently, to resolve the feature degradation caused by amplified noise and feature inconsistencies, we devise a purification strategy with next two modules. 
At the signal level, we design a Dual-Branch Wavelet Enhanced Denoiser (\textbf{WTDen}) module, which corrects global distortions with a Wavelet Mamba and remedies local inconsistencies via Wavelet Convolution. Subsequently, at the semantic level, we design a Adaptive Feature Selector (\textbf{AdpSel}) module to refine collaborative features by selecting critical regions and pruning artifacts, ensuring the robustness of the final fusion.
% At the signal level, the Dual-Branch Wavelet Enhanced Denoiser (\textbf{WTDen}) corrects global distortions with a Wavelet Mamba and remedies local inconsistencies via Wavelet Convolution. Subsequently, at the semantic level, the Adaptive Feature Selector (\textbf{AdpSel}) refines collaborative features by selecting critical regions and pruning artifacts, ensuring the robustness of the final fusion.
% The first stage addresses noise at a signal level with the proposed Dual-Branch Wavelet Enhanced Denoiser (\textbf{WTDen}). It leverages the wavelet domain to reconstruct features: a Wavelet Mamba corrects global distortions while Wavelet Convolution remedies local inconsistencies. In the second stage, the collaboration is refined at a semantic level with our Adaptive Feature Selector (\textbf{AdpSel}). It operates by selecting the most critical feature regions while pruning inconsistent artifacts, thereby ensuring a robust final synthesis.

To validate the effectiveness of CATNet, we conducted extensive experiments on various cooperative perception datasets. Using the V2XSet as an example, results demonstrate that our method achieves an average AP@0.5/AP@0.7 improvement of 5.7\%/2.5\% compared to the second-best approach, and outperforms the single-vehicle baseline by 16.0\%/12.7\% under noisy and latency scenarios. 
Our contributions can be summarized as follows.

\begin{itemize}
    \item We propose CATNet, a novel cooperative perception framework specifically engineered to overcome two fundamental challenges in multi-agent systems: communication asynchrony and feature inconsistency.

    \item We design STSync for robust temporal alignment and  introduce a dual-purification strategy that uses WTDen for signal denoising and AdpSel for semantic refinement.
    
    \item We conduct extensive experiments on large-scale datasets, demonstrating that CATNet achieves state-of-the-art performance and shows remarkable robustness, especially under severe communication delays and noise.
\end{itemize}

\begin{figure*}[!t] 
    \centering 
    \includegraphics[width=0.9\textwidth]{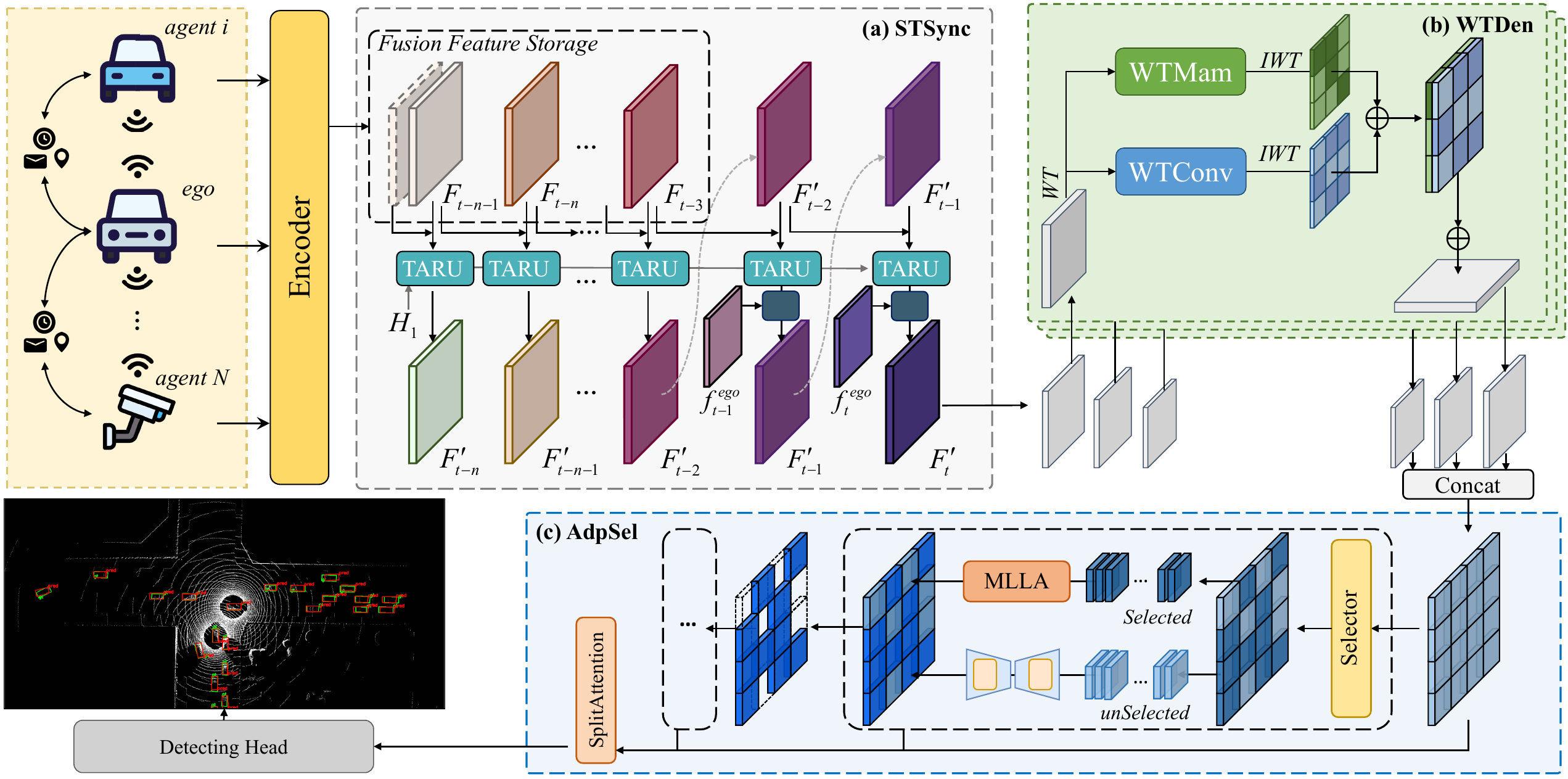} 
    \caption{Architectural overview of the CATNet framework. The ego vehicle maintains a fusion feature bank that processes multi-agent transmitted features, employing temporal prediction mechanisms for current-time feature estimation. These refined features subsequently undergo denoising operations and salient feature extraction, achieving robust cross-agent feature fusion via learnable aggregation weights.} 
    \label{fig:overall} 
\end{figure*}
\section{Related Work}
\label{sec:related}

% \subsection{Cooperative Perception} 
\noindent \textbf{Cooperative Perception.}
Cooperative perception technology enhances the environmental sensing capabilities \cite{xu2022opv2v, wang2020v2vnet, dairv2x} of individual agents \cite{zhu2020deformable, zhao2024ms} through information sharing among multiple agents, and has become a research hotspot in autonomous driving and multi-robot systems. Mainstream research methods are divided into three categories. Early fusion \cite{cooper, arnold2022cooperative} integrates information at the raw data level, achieving global perception by sharing point cloud or image data, but it imposes stringent requirements on communication bandwidth. Late fusion \cite{su2023uncertainty, dairv2x} occurs at the individual decision-making level, but it is limited by the local perception capabilities of individual agents and suffers from significant information loss, making it difficult to handle occlusions or distant targets. Intermediate fusion \cite{xu2022opv2v, xu2022v2xvit, yang2023how2comm} facilitates information exchange at the feature level, enabling efficient collaboration through the sharing of intermediate feature maps. This approach has become the most prominent technical paradigm in the field, and many studies have achieved promising results in terms of perception accuracy. However, current intermediate fusion methods remain vulnerable in complex environments with heterogeneous noise profiles and dynamic latency. 

% \subsection{Multi-Agent Latency Mitigation} 
\noindent \textbf{Multi-Agent Latency Mitigation.}
Cooperative perception systems face asynchronous feature misalignment during fusion due to transmission delays and communication interruptions \cite{gutwin2004revealing}. Current solutions primarily adopt two strategies: Methods like V2X-ViT \cite{xu2022v2xvit} and V2VNet \cite{wang2020v2vnet} concatenate asynchronous features and implicitly learn spatio-temporal correlations through deep networks, but fail to capture dynamic scene evolution, while alignment-focused approaches such as SyncNet \cite{syncnet} (using P-LSTM for timestamp synchronization) and MRCNet \cite{mrcnet} (employing feature prediction networks) attempt local compensation but lack global spatio-temporal coherence modeling across agents. These limitations lead to incomplete predictions and iterative error accumulation, as neither paradigm effectively addresses multi-agent temporal consistency or prevents error propagation through collaborative processing chains.

% \subsection{Multi-Source Noise Suppression} 
\noindent \textbf{Multi-Source Noise Suppression.}
Cooperative perception systems suffer from multi-source noise, including sensing inaccuracies, communication distortions, model deviations, and latency artifacts across the pipeline \cite{chen2025cora, chen2026coopdiffdiffusionguidedapproachcooperation}. Current solutions primarily adopt two paradigms: knowledge distillation and graph optimization. The first approach, represented by DiscoNet \cite{disconet} and DI-V2X \cite{li2024di}, employs teacher-student frameworks to suppress interference through clean feature guidance, yet exhibits critical limitations including teacher model dependency and inadequate performance in compounded interference scenarios. The second paradigm exemplified by CoAlign \cite{coalign} utilizes graph-based spatial alignment to correct sensor distortions, but suffers from temporal inconsistency in dynamic environments where communication delays induce asynchronous feature misalignment and cascading errors. 
However, a key limitation persists: existing methods primarily purify features at a signal level, while overlooking residual semantic inconsistencies and contextual artifacts. This incomplete approach to purification critically undermines system reliability.

\section{Methods}

\subsection{Overall Architecture}
To address the dual challenges of non-deterministic transmission delays and multi-source noise interference in cooperative perception scenarios, we propose a Collaborative Alignment and Transformation Network (CATNet). The complete workflow comprises four components: a feature encoder, feature transmission, CATNet, and a feature decoder. Our framework operates through the following pipeline:
\begin{equation}
F^t_i = f_{\text{encoder}}\left(X_i^t\right), \quad i \in N  \\
\end{equation}
\begin{equation}
\hat{F}_{i \to \text{ego}}^{t-\tau} = \xi_{i \to \text{ego}}^{t-\tau}\left(F_i^{t-\tau}\right) 
\end{equation}
\begin{equation}
\widetilde{F}_{\text{fused}}^t = \text{CATNet}\Bigl( \left\{ \hat{F}_{i \to \text{ego}}^{t'} \right\}_{t' \le t-\tau}, \left\{ F_{\text{ego}}^{t'} \right\}_{t' \le t} \Bigr) \\
\end{equation}
\begin{equation}
\widetilde{Y}_{\text{fused}}^t = f_{\text{decoder}}\left(\widetilde{F}_{\text{fused}}^t\right)
\end{equation}

In this framework, each agent $i$ first encodes sensor data \( X_i^t \) into features \( F_i^t \) via an encoder \( f_{\text{encoder}} \). After transmission, these features arrive at the ego vehicle with a delay $\tau$ and are transformed via coordinate transformation $\xi_{i \to \text{ego}}^{t-\tau}$ to its coordinate frame, resulting in $\hat{F}_{i \to \text{ego}}^{t-\tau}$. The set of all such received features at each timestep is denoted as ${F}_{\text{fused}}^{t-\tau}$. As illustrated in Figure \ref{fig:overall}, CATNet then fuses these aligned features with the ego's local historical features through three modules. Specifically, it utilizes a temporal alignment module to resolve asynchrony and a dual-stage denoising mechanism to suppress distortions, thereby producing fused features $\widetilde{F}_{\text{fused}}^t$. Finally, a decoder $f_{\text{decoder}}$ generates perception outputs $\widetilde{Y}_{\text{fused}}^t$.

\subsection{Spatio-Temporal Recurrent Synchronization}
\begin{figure}[!t] 
    \centering % 使图片居中显示
    \includegraphics[width=0.45\textwidth]{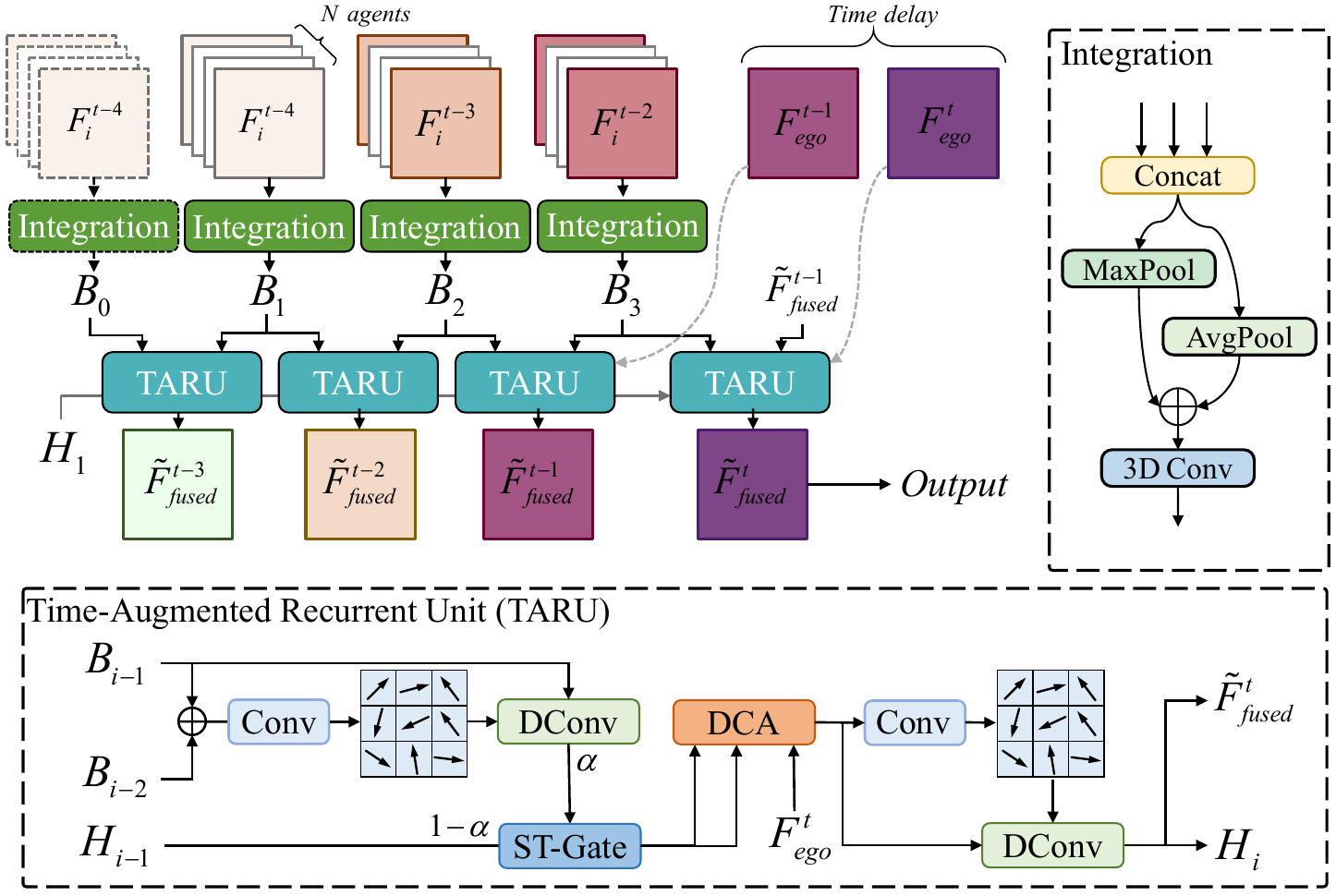}
    \caption{Architecture of the proposed STSync module. The module employs adaptive integration for preliminary multi-agent feature fusion, coupled with TARU that incorporates ego-vehicle features to achieve temporally coherent feature prediction.} % 为图片添加标题
    \label{fig:module1} % 为图片添加标签，方便引用
\end{figure}

Spatio-Temporal Recurrent Synchronization (STSync) is designed to address latency alignment in multi-agent cooperative perception. As shown in Figure \ref{fig:module1}, it employs a dynamic timestamp calibration mechanism to sequentially model inter-frame motion. This unique approach compensates for latency-induced feature misalignment, ultimately ensuring robust spatio-temporal coherence across asynchronous data streams.

For a scene with $N$ agents, the ego vehicle receives delayed features $F^{t-\tau}_{\text{agents}}\in \mathbb{R}^{N\times C\times H\times W}$. These are first processed by a dual-stream multi-scale fusion module (Integration) to generate a unified representation:
\begin{equation}
F_{\text{fused}}^{t-\tau} = \mathcal{C}_3\left( \mathrm{Concat}\left[\mathcal{M}_p(F^{t-\tau}_{\text{agents}}),\, \mathcal{A}_p(F^{t-\tau}_{\text{agents}})\right] \right)
\end{equation}
where $\mathcal{C}_3$ denotes a 3D convolution, $\mathcal{M}_p$ and $\mathcal{A}_p$ represent global max and average pooling operations, respectively.

The ego vehicle maintains a feature buffer, $\mathcal{B}$, which stores the $K$ most recently fused historical features. This sequence is represented as $\mathcal{B} = (B_1, B_2, \ldots, B_K)$, where each element $B_j$ corresponds to a feature map from a historical timestep, specifically $F_{\text{fused}}^{t-\tau-K+j}$.

% 如图三所示  两个初始化 表述不清
To model the temporal dependencies within this feature buffer, we introduce the Time-Augmented Recurrent Unit (TARU). Its hidden state is initialized with the earliest feature, $H_1 = B_1$. The module then iteratively operates for each timestep $i$ from $2$ to $K$, using a zero-tensor $B_0$ to handle the initial boundary condition. The operations at each step $i$ are:

\textbf{Motion Prediction.} To capture temporal dynamics, a motion offset is predicted by observing the two preceding features.
\begin{equation}
\Delta B_i = \text{Conv}\left(\text{Concat}\big[B_{i-2}, B_{i-1}\big]\right)
\end{equation}

\textbf{Feature Warping.} The offset is used to warp the previous feature via deformable convolution, producing a motion-aligned feature $\hat{B}_i$.
    \begin{equation}
    \hat{B}_i = \text{DeformConv}\left(B_{i-1}, \Delta B_i\right)
    \end{equation}

% 附录 ST-Gate
\textbf{State Fusion.} The aligned feature $\hat{B}_i$ is fused with the previous hidden state $H_{i-1}$ using the Spatio-Temporal Gate (ST-Gate). ST-Gate employs parallel spatial and channel attention mechanisms (detailed in \textit{Appendix.1}) to ensure an adaptive fusion. These pathways process the concatenated hidden state and warped feature, computing an adaptive gating coefficient $\alpha_i$. This coefficient dynamically balances the contribution of historical context ($H_{i-1}$) and current motion-aligned information ($\hat{B}_i$). The fusion is then performed as a weighted sum:
    \begin{equation}
    S_i = (1 - \alpha_i) \cdot H_{i-1} + \alpha_i \cdot \hat{B}_i
    \end{equation}
This method enables the model to intelligently weigh temporal memory against current observations.

\textbf{Feature Update.} This intermediate state $S_i$ undergoes further refinement to update the hidden state $H_i$. The final hidden state $H_K$ serves as the predicted feature.
    \begin{equation}
    H_i = \text{DeformConv}\left(\text{Conv}\left(S_i\right), S_i\right), \quad  \widetilde{F}_{\text{fused}}^t = H_K
    \end{equation}

Finally, this predicted feature $\widetilde{F}_{\text{fused}}^t$ is refined using Deformable Cross-Attention (DCA) with the ego-vehicle's own real-time feature $F_{\text{ego}}^t$ as a spatial prior. This crucial step anchors the temporally-predicted features to the spatially accurate reality of the ego-vehicle, ensuring robust spatio-temporal alignment.

\subsection{Dual-Branch Wavelet Enhanced Denoiser}

\begin{figure}[!t] 
    \centering % 使图片居中显示
    \includegraphics[width=0.45\textwidth]{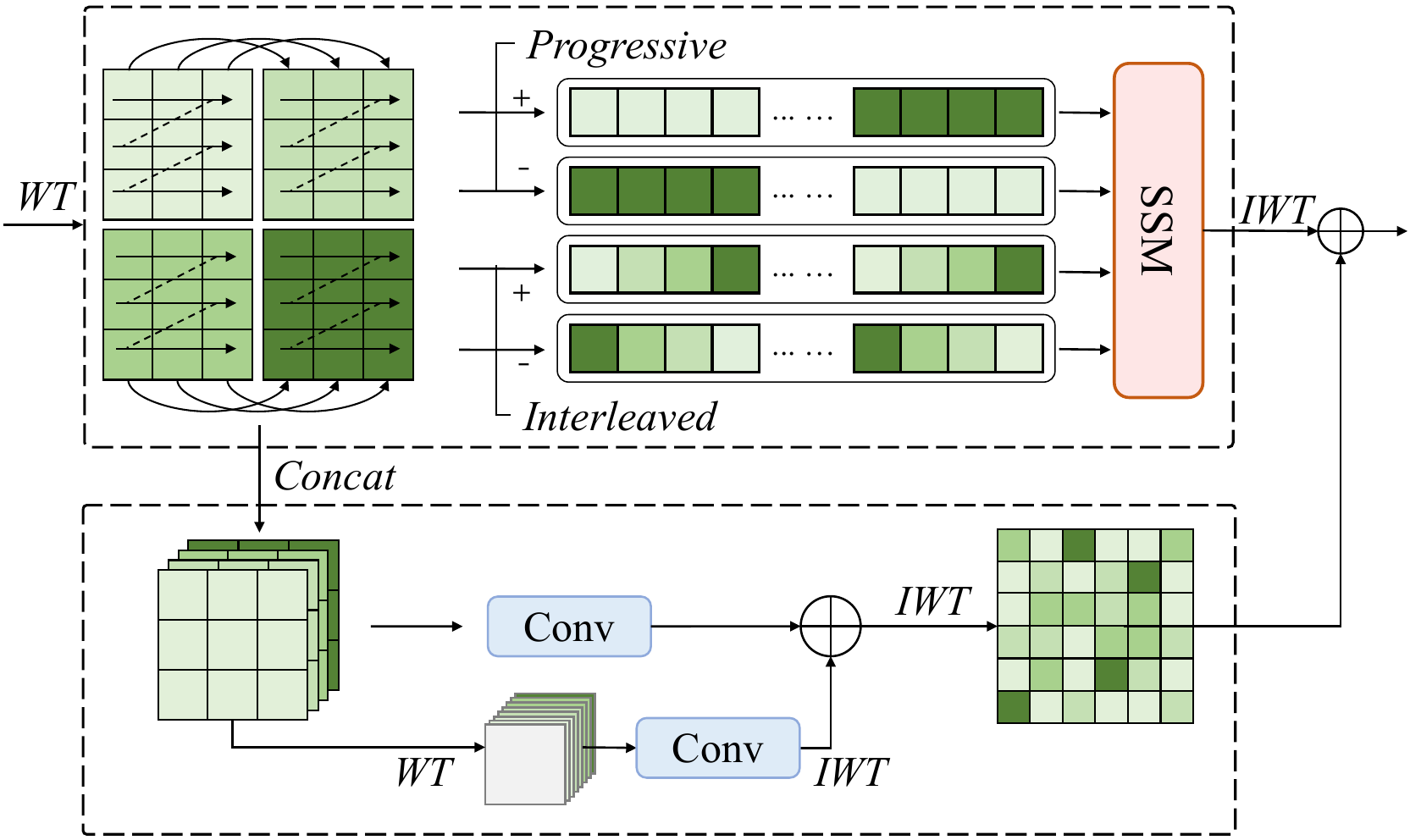}
    \caption{Architecture of the WTDen module. The module performs global and local feature denoising via a dual-branch design integrating wavelet-Mamba and wavelet convolution.} % 为图片添加标题
    \label{fig:module2} % 为图片添加标签，方便引用
\end{figure}

% 小波变换的具体事宜 后面补一个见附录
% Despite STSync's effectiveness in synchronizing multi-source features, two subsequent challenges persist: First, its iterative temporal processing can amplify high-frequency feature noise. Second, resolving inherent feature inconsistencies between the ego vehicle and collaborators is critical for robust alignment. Inspired by the prior works \cite{syncnet, mrcnet}, we introduce the Dual-Branch Wavelet Enhanced Denoiser (WTDen) to address both issues concurrently. Operating in the wavelet domain, WTDen inherently isolates and suppresses noise. Moreover, its dual-branch architecture explicitly enforces feature alignment on global and local scales.
Despite its effectiveness in feature synchronization, STSync still faces a persistent issue with residual artifacts, which primarily occur at the signal level: its iterative temporal processing can amplify high-frequency feature noise, and inherent inconsistencies between agents can corrupt local feature structures. To address these signal-level distortions, we introduce the Dual-Branch Wavelet Enhanced Denoiser (WTDen) as the first stage of our purification pipeline (Figure \ref{fig:module2}). WTDen operates in the wavelet domain, which allows it to isolate and suppress noise. Moreover, its dual-branch architecture enforces feature alignment on global and local scales.

The module first decomposes the fused feature map \(\widetilde{F}_{\text{fused}}^t \in \mathbb{R}^{C \times {H} \times {W}}\) using the 2D Haar Wavelet Transform (WT). This process separates the spatial features into four subbands, each of dimension \(\in C \times \frac{H}{2} \times \frac{W}{2}\).  
\begin{equation}
F_{LL}, F_{LH}, F_{HL}, F_{HH} = \text{WT}\left(\widetilde{F}_{\text{fused}}^t\right)
\end{equation}

$F_{LL}$ captures the low-frequency structural information, while $ F_{LH}, F_{HL}, F_{HH}$ encode the high-frequency details.

%SSM扫描顺序 说明
\textbf{Wavelet Mamba.} This branch is designed to capture long-range spatial relationships and correct global feature misalignment among agents. We employ a dual-path progressive fusion strategy within the Wavelet Mamba. One path is a forward integration that sequentially processes subbands from high-to-low frequency ($F_{HH} \rightarrow F_{LL}$). This prioritizes high-frequency edge information to effectively compensate for detail loss. Concurrently, an interleaved scanning strategy processes features from all four subbands at each spatial location to capture cross-band correlations. Detailed illustration of scanning strategies is available in \textit{Appendix.1}. The dual-path design incorporates reverse processes for both integration and scanning, ensuring comprehensive feature aggregation across all directions and scales. The outputs from four scanning paths ($L_{\text{prog}}^{\pm}, L_{\text{inter}}^{\pm}$) are aggregated through SSM, and the resulting feature is processed to yield the four enhanced subbands: $\{F_{\text{LL}}^{\text{fused}}, F_{\text{LH}}^{\text{fused}}, F_{\text{HL}}^{\text{fused}}, F_{\text{HH}}^{\text{fused}}\}$. These are then fused via Inverse Wavelet Transform (IWT) to restore a globally aligned feature:
\begin{equation}
F_{\text{mam}} = \text{IWT}\left( \left\{ F_{\text{LL}}^{\text{fused}}, F_{\text{LH}}^{\text{fused}}, F_{\text{HL}}^{\text{fused}}, F_{\text{HH}}^{\text{fused}} \right\} \right)
\end{equation}

\textbf{Wavelet Convolution.} To complement the global alignment, this branch addresses local feature degradation and inconsistencies. It models fine-grained local patterns, ensuring that the features for each vehicle are coherent and free from localized noise. The four wavelet subbands are first concatenated, creating a tensor \(F_{\text{wt}} \in \mathbb{R}^{4C \times \frac{H}{2} \times \frac{W}{2}}\). A hierarchical filtering process is then applied: 
\begin{equation}
  F_{\text{conv}} = \text{IWT}\left( \text{IWT}\left( \text{Conv}\left( \text{WT}\left( F_{\text{wt}} \right) \right) \right) \oplus \text{Conv}\left( F_{\text{wt}} \right) \right)
\end{equation}

The final output combines both branch features:  
\begin{equation}
  F_{\text{denoise}} = F_{\text{mam}} + F_{\text{conv}}
\end{equation}

\subsection{Adaptive Feature Selector} 

% The Adaptive Feature Selector (AdpSel) module is designed for efficient and robust fusion of multi-agent perception features. Unlike conventional methods that treat all feature regions equally \cite{xu2022v2xvit, zhang2024ermvp}, AdpSel pioneers a hierarchical selection mechanism that prioritizes salient, cross-vehicle features. This shifts the fusion paradigm from simple data aggregation to a targeted synthesis of critical information, yielding a more focused and powerful feature representation.
The Adaptive Feature Selector (AdpSel) constitutes the second stage of synergistic purification, specifically designed to address higher-order semantic artifacts that signal-level filters cannot remove. By redefining saliency as a proxy for semantic coherence, AdpSel performs context-aware synthesis to selectively enhance coherent feature regions, thereby filtering out semantic noise.

\begin{figure}[!t]
    \centering % 使图片居中显示
    \includegraphics[width=0.48\textwidth]{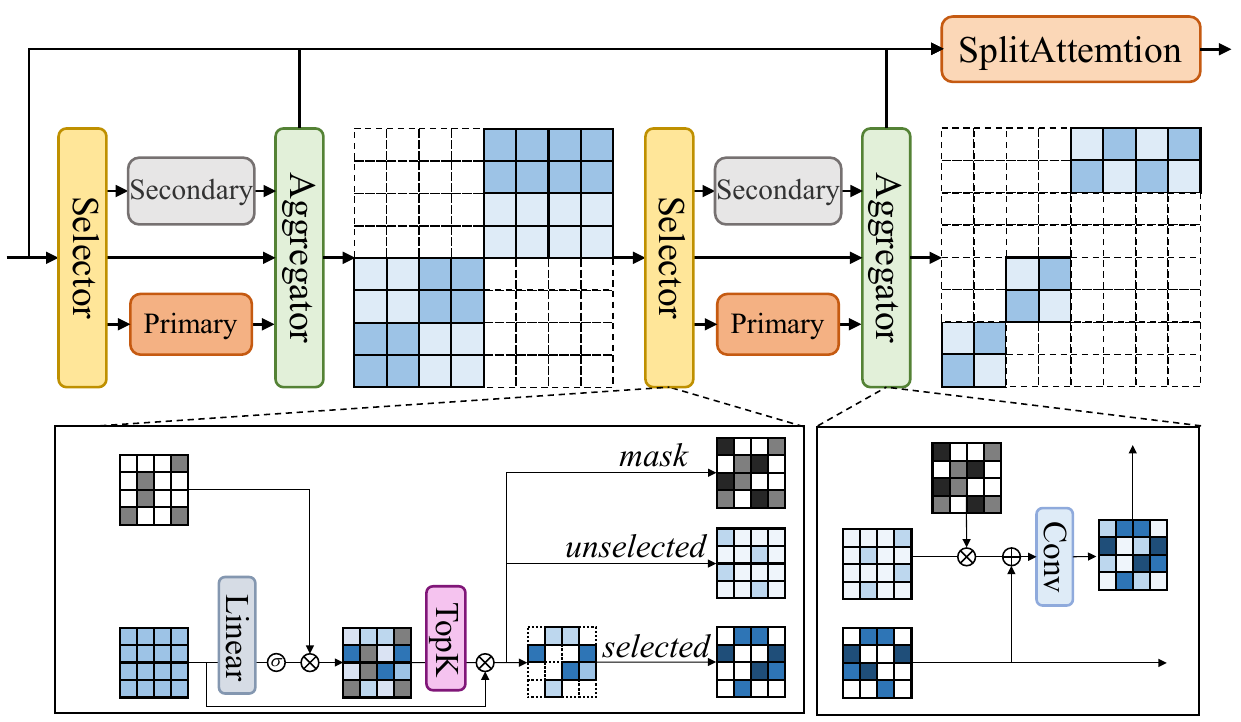}
    \caption{Architecture of the AdpSel module. The module divides features through a Selector, applies the MLLA module to enhance critical features, and employs inverted bottleneck convolution for feature compensation in non-critical regions. For critical features, these operations are iteratively applied, and the results are fused.} % 为图片添加标题
    \label{fig:module3} % 为图片添加标签，方便引用
\end{figure}

AdpSel operates iteratively across a set of predefined window scales $\{S_1, S_2, \dots, S_n\}$. As shown in Figure \ref{fig:module3}, the input feature map $F_{\text{denoise}} \in \mathbb{R}^{N \times C \times H \times W}$ undergoes three key steps at each scale $S_i$,:

\textbf{Coherence-Aware Block Selection}. The feature map is first partitioned into non-overlapping blocks $F_{S_i}$. A lightweight linear selector $\phi(\cdot)$ assigns an importance score to each block, producing a score map $\Phi_{S_i}$. Based on these scores, we identify the top-$k\%$ of blocks as selected features ($F_{S_i}^{\text{selected}}$) and the remainder as unselected features ($F_{S_i}^{\text{unselected}}$). Let $M_{S_i}^{\text{topk}} = \text{TopK}(\Phi_{S_i})$ be the binary mask for the top-k blocks, then:
\begin{equation}
    F_{S_i}^{\text{selected}} = M_{S_i}^{\text{topk}} \odot F_{S_i}
\end{equation}
\begin{equation}
    F_{S_i}^{\text{unselected}} = (1 - M_{S_i}^{\text{topk}}) \odot F_{S_i}
\end{equation}

% We first define a set of window sizes $\{S_1, S_2, \dots, S_n\}$ for feature partitioning and an initial mask $\text{mask}_{S_i}$. As shown in Figure \ref{fig:module3}, the input feature \(F_{\text{denoise}} \in \mathbb{R}^{N \times C \times H \times W}\) undergoes sequential processing: partitioned into localized blocks \(F_{S_i} \in \mathbb{R}^{N \times C \times \frac{H}{S_i} \times \frac{W}{S_i} \times (S_i \times S_i)}\) using predefined window sizes, evaluated by a fully-connected selector \(\phi(\cdot)\) for block-wise importance scoring, dynamically suppressing low-response regions via \(\text{mask}_{S_i}\) while preserving high-saliency blocks with maintained dimensional integrity for subsequent fusion.

% \begin{equation}
% \Phi_{S_i} = \phi(F_{S_i}) \odot \text{mask}_{S_i}, \quad \Phi_{S_i} \in \mathbb{R}^{N \times \frac{H}{S_i} \times \frac{W}{S_i}}
% \end{equation}

% The remaining tokens are then divided into the top $k\%$ important tokens $F_{S_i}^{\text{selected}}$ and non-important tokens $F_{S_i}^{\text{unselected}}$.

% \begin{equation}
% F_{S_i}^{\text{selected}}, F_{S_i}^{\text{unselected}} = \text{TopK}(\Phi_{S_i}) \odot F_{S_i},  (1 - \text{TopK}(\Phi_{S_i})) \odot F_{S_i}
% \end{equation}

% The mask is hierarchically updated by aggregating discarded regions from the current scale through the operation \( \text{mask}_{S_{i+1}} = \text{UpSample}(\text{mask}_{S_i} \odot (1 - \text{TopK}(\Phi_{S_i}))) \), where discarded low-salience regions (identified via \( 1 - \text{TopK}(\Phi_{S_i}) \)) are upsampled and fused with the original mask to guide subsequent selections. This cross-scale propagation ensures progressive refinement, allowing the module to iteratively focus on critical regions while maintaining a coherent attention pathway from coarse-to-fine granularity.

\textbf{Hierarchical Mask Refinement.} A key innovation of AdpSel is its cross-scale mask propagation. Low-saliency regions discarded at a finer scale $S_i$ are used to refine the selection mask for the next, coarser scale $S_{i+1}$. Specifically, the mask of discarded regions $(1 - M_{S_i}^{\text{topk}})$ is upsampled and integrated into the initial mask for the subsequent stage. This ensures that the model progressively focuses on globally salient areas while avoiding redundant computation on regions already identified as unimportant.
\begin{equation}
    \text{mask}_{S_{i+1}} = \text{mask}_{\text{initial}} - \text{UpSample}(1 - M_{S_i}^{\text{topk}})
\end{equation}

\begin{table*}[t]
\caption{Comparative evaluation of detection accuracy and model parameters with state-of-the-art methods on OPV2V, V2XSet, and DAIR-V2X datasets under latency and noise conditions. AP@0.5 and AP@0.7 metrics are presented separately. $^*$Regular methods. $^\dagger$Noise-robust methods. $^\ddagger$Latency-aware methods.}
\label{tab:comparison}
\centering
% 使用标准列类型 'c'，因为 AAAI 对 siunitx 的 'S' 列类型可能视为改变外观而不允许
% 如果您之前在 preamble 中使用了 \usepackage{tabularx}，请务必删除或注释掉该行。
\begin{tabular}{@{} l l c *{6}{c} @{}} % 将 S 列类型改为 c
\hline
\multirow{2}{*}{Method} & \multirow{2}{*}{Published} & {\multirow{2}{*}{Params}} & \multicolumn{2}{c}{OPV2V} & \multicolumn{2}{c}{V2XSet} & \multicolumn{2}{c}{DAIR-V2X} \\
\cmidrule(lr){4-5} \cmidrule(lr){6-7} \cmidrule(lr){8-9}
& & & {AP@0.5} & {AP@0.7} & {AP@0.5} & {AP@0.7} & {AP@0.5} & {AP@0.7} \\
\hline
No Fusion & {--} & {--} & 0.738 & 0.509 & 0.698 & 0.516 & 0.625 & 0.446 \\
\hline
$^*$Where2comm \cite{hu2022where2comm} & {NIPS'22} & 11.43 M & 0.699 & 0.402 & 0.757 & 0.495 & 0.654 & 0.502 \\
$^*$CoMamba \cite{li2024comamba} & {Arxiv'24} & {\textbf{09.15} M} & 0.743 & 0.558 & 0.726 & 0.531 & 0.635 & 0.479 \\
\hline
$^\dagger$V2X-ViT \cite{xu2022v2xvit} & {ICRA'22} & 13.50 M & 0.817 & 0.633 & 0.797 & 0.593 & 0.696 & 0.517 \\
$^\dagger$CORE \cite{wang2023core} & {ICCV'23} & {09.19 M} & 0.831 & 0.628 & 0.764 & 0.504 & \multicolumn{2}{c}{--} \\
$^\dagger$CoAlign \cite{coalign} & {ICRA'23} & {11.43 M} & 0.713 & 0.526 & 0.786 & 0.581 & 0.693 & 0.549 \\
$^\dagger$DSRC \cite{zhang2024dsrc} & {AAAI'25} & {40.64 M} & 0.789 & 0.653 & 0.801 & 0.596 & 0.702 & 0.559 \\
\hline
$^\ddagger$How2comm \cite{yang2023how2comm} & {NIPS'23} & 35.80 M & 0.739 & 0.562 & 0.749 & 0.624 & 0.624 & 0.472 \\
$^\ddagger$ERMVP \cite{zhang2024ermvp} & {CVPR'24} & {12.42 M} & 0.820 & 0.679 & 0.744 & 0.499 & 0.674 & 0.554 \\
$^\ddagger$MRCNet \cite{mrcnet} & {CVPR'24} & {19.71 M} & 0.814 & 0.617 & 0.817 & 0.618 & 0.665 & 0.539 \\
\hline
\textbf{CATNet} & {--} & {09.95 M} & \bfseries 0.843 & \bfseries 0.686 & \bfseries 0.858 & \bfseries 0.643 & \bfseries 0.723 & \bfseries 0.565 \\
\hline
\end{tabular}
\end{table*}

\textbf{Dual-Path Feature Enhancement.} To maximize robustness, features are processed through a dual-path architecture based on their semantic confidence score. The high-coherence selected blocks ($F_{S_i}^{\text{selected}}$) are processed by the MLLA module \cite{mlla} to capture complex contexts.
\begin{equation}
    F_{S_i}^{\text{enhanced}} = \text{MLLA}\left(F_{S_i}^{\text{selected}}\right)
\end{equation}
Concurrently, the unselected blocks ($F_{S_i}^{\text{unselected}}$) are processed by a lightweight Inverted Bottleneck (IB) layer \cite{ib} to efficiently recover supplementary information with minimal overhead.
\begin{equation}
    F_{S_i}^{\text{recovered}} = \text{IB}\left(F_{S_i}^{\text{unselected}}\right)
\end{equation}

% For the selected high-weight salient feature blocks $F_{S_i}^{\text{selected}}$, we further enhance their importance using the Mamba-Like Linear Attention (MLLA) \cite{mlla} module. Meanwhile, for $F_{S_i}^{\text{unselected}}$, we employ a lightweight Inverted Bottleneck (IB) \cite{ib} layer to compensate for information, preserving potentially useful features:

% \begin{equation}
% F_{S_i}^{\text{enhanced}} = \text{MLLA}\left(F_{S_i}^{\text{selected}}\right)
% \end{equation}
% \begin{equation}
% F_{S_i}^{\text{recovered}} = \text{IB}\left(F_{S_i}^{\text{unselected}}\right)
% \end{equation}

Finally, at each scale, an Aggregator module fuses the enhanced and recovered features via simple convolutions to produce a scale-specific output $F_{S_i}^{\text{fused}}$. After iterating through all scales, the resulting feature set $\{F_{S_1}^{\text{fused}}, \dots, F_{S_n}^{\text{fused}}\}$ is fused by a SplitAttention \cite{xu2022v2xvit} layer to generate the final output $F_{\text{out}}$.
\begin{equation}
    F_{S_i}^{\text{fused}} = \text{Aggregator}\left(F_{S_i}^{\text{enhanced}}, F_{S_i}^{\text{recovered}}\right)
\end{equation}
\begin{equation}
    F_{\text{out}} = \text{SplitAttention}\left(F_{S_1}^{\text{fused}}, F_{S_2}^{\text{fused}}, \dots, F_{S_n}^{\text{fused}}\right)
\end{equation}

% Next, an Aggregator composed of simple convolutions is used to fuse the enhanced features and compensated features together. It then outputs the fused result at scale $S_i$ after the convolution.

% \begin{equation}
% F_{S_i}^{\text{fused}} = \text{Aggregator}\left(F_{S_i}^{\text{enhanced}}, F_{S_i}^{\text{recovered}}, \text{mask}_{S_i}\right)
% \end{equation}

% Building upon $F_{S_i}$, the above operations are repeated for all window sizes, resulting in an output set $\{F_{S_1}, F_{S_2}, \dots, F_{S_n}\}$. These results are then fused with SplitAttention \cite{xu2022v2xvit} as the final output:

% \begin{equation}
% F_{\text{out}} = \text{SplitAttenion}\left(F_{S_1}^{\text{fused}}, F_{S_2}^{\text{fused}}, \dots, F_{S_n}^{\text{fused}}\right)
% \end{equation}

\begin{figure*}[!t] % 使用 figure* 环境插入跨栏图片，并指定优先放在页面顶部
    \centering % 使图片居中显示
    \includegraphics[width=1.0\textwidth]{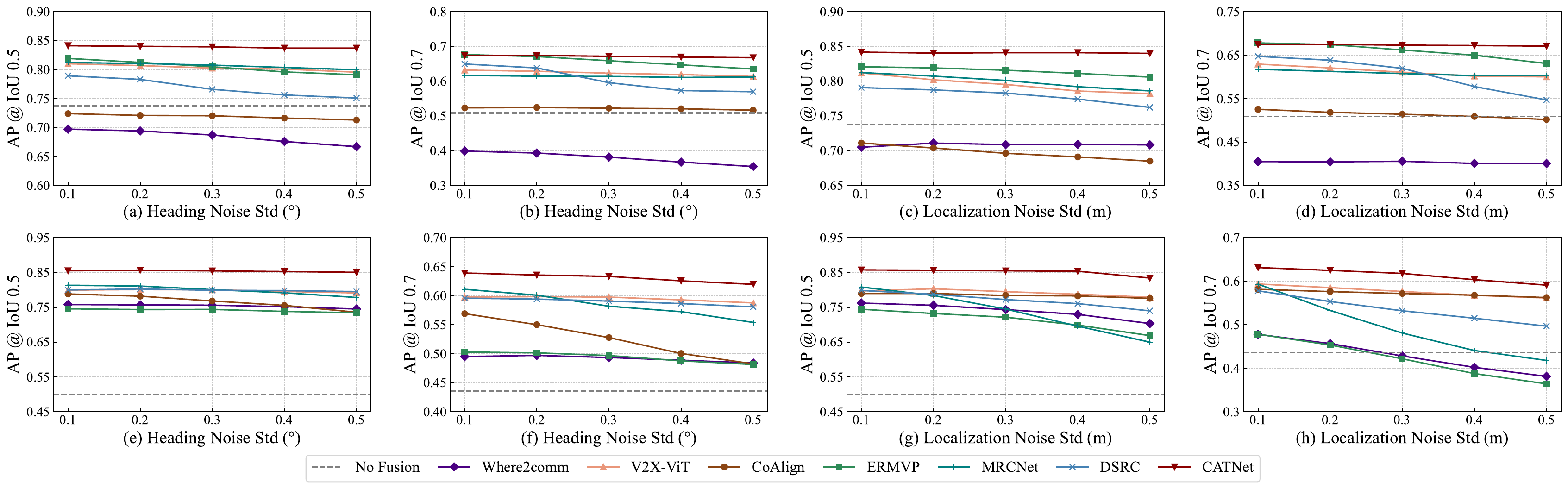} % 插入图片
    \caption{Quantitative analysis of noise robustness measured by AP@0.5/0.7: across OPV2V and V2XSet datasets; upper panels depict OPV2V performance, lower panels show V2XSet performance.} % 为图片添加标题
    \label{fig:combined_noise} % 为图片添加标签，方便引用
\end{figure*}

\section{Experiments}

\subsection{Experimental Setup}

\noindent \textbf{Datasets and Metrics.} We comprehensively evaluate our proposed CATNet on three large-scale public benchmarks for cooperative perception: OPV2V \cite{xu2022opv2v}, the first large-scale vehicle-to-vehicle dataset; V2XSet \cite{xu2022v2xvit}, the first vehicle-to-infrastructure dataset; and DAIR-V2X \cite{dairv2x}, a real-world dataset that captures complex and challenging scenarios. For a fair comparison, all methods use the standard PointPillar \cite{lang2019pointpillars} backbone and are evaluated using the official Average Precision (AP) at IoU of 0.5 and 0.7.

\noindent \textbf{Comparison Methods.} We compare CATNet against a range of state-of-the-art approaches, categorized by their primary strength: regular methods (e.g., Where2comm \cite{hu2022where2comm}), noise-robust methods (e.g., V2X-ViT \cite{xu2022v2xvit}, DSRC \cite{zhang2024dsrc}), and latency-aware methods (e.g., ERMVP \cite{zhang2024ermvp}, MRCNet \cite{mrcnet}). We choose the naive intermediate method as the baseline. Further details are provided in \textit{Appendix.1}.

\begin{table*}[t]
\caption{Evaluation of methods under varying delay conditions on the OPV2V dataset.}
\centering
% 使用标准列类型 'c'，因为 AAAI 对 siunitx 的 'S' 列类型可能视为改变外观而不允许
\begin{tabular}{@{} l *{10}{c} @{}} % 将 S 列类型改为 c
\hline
Method & \multicolumn{2}{c}{0ms} & \multicolumn{2}{c}{0-200ms} & \multicolumn{2}{c}{0-300ms} & \multicolumn{2}{c}{0-400ms} & \multicolumn{2}{c}{0-500ms} \\
\cmidrule(lr){2-3} \cmidrule(lr){4-5} \cmidrule(lr){6-7} \cmidrule(lr){8-9} \cmidrule(lr){10-11}
& AP@0.5 & AP@0.7 & AP@0.5 & AP@0.7 & AP@0.5 & AP@0.7 & AP@0.5 & AP@0.7 & AP@0.5 & AP@0.7 \\
\hline
CoAlign & 0.881 & 0.748 & 0.709 & 0.517 & 0.658 & 0.524 & 0.631 & 0.508 & 0.610 & 0.495 \\
V2X-ViT & 0.863 & 0.724 & 0.774 & 0.580 & 0.738 & 0.557 & 0.716 & 0.538 & 0.697 & 0.530 \\
ERMVP & 0.831 & 0.686 & 0.824 & 0.643 & 0.783 & 0.630 & 0.767 & 0.613 & 0.745 & 0.600 \\
MRCNet & 0.848 & 0.715 & 0.836 & 0.647 & 0.774 & 0.638 & 0.759 & 0.625 & 0.738 & 0.615 \\
DSRC & 0.887 & 0.752 & 0.841 & 0.598 & 0.731 & 0.552 & 0.709 & 0.522 & 0.640 & 0.508 \\
CATNet & \textbf{0.896} & \textbf{0.763} & \textbf{0.856} & \textbf{0.673} & \textbf{0.806} & \textbf{0.657} & \textbf{0.774} & \textbf{0.638} & \textbf{0.756} & \textbf{0.624} \\
\hline
\end{tabular}
\label{tab:latency_evaluation}

\end{table*}

\subsection{Quantitative Results} 

% \noindent \textbf{Detection Performance Analysis.} As quantified in Table \ref{tab:comparison}, our method achieves consistent superiority across all evaluation metrics. Specifically, on the OPV2V dataset, our model achieves improvements of 1.2\%/3.3\% in the key metrics of $AP@0.5/AP@0.7$, respectively, compared to the second-best model. On the V2XSet dataset, the performance gap is even more pronounced, with improvements of 5.7\%/2.5\% in $AP@0.5/AP@0.7$. Similarly, on the DAIR-V2X dataset, our model outperforms the second-best model by 2.7\%/1.1\% in these metrics. These results demonstrate that our model can achieve more accurate multi-vehicle object detection in complex vehicular environments.

\noindent \textbf{Detection Performance Analysis.} As quantified in Table \ref{tab:comparison}, our method (CATNet) achieves consistent superiority across all evaluation metrics. Specifically, on the OPV2V dataset, our model achieves improvements of 1.2\% / 0.7\% in the key metrics of $AP@0.5 / AP@0.7$, respectively, compared to the second-best model. On the V2XSet dataset, the performance gap is even more pronounced, with improvements of 4.1\% / 1.9\% in $AP@0.5 / AP@0.7$. Similarly, on the DAIR-V2X dataset, our model outperforms the second-best model by 2.1\% / 0.6\% in these metrics. These results demonstrate that our model can achieve more accurate multi-vehicle object detection in complex vehicular environments.

\noindent \textbf{Noise Robustness Evaluation.}  Figure \ref{fig:combined_noise} systematically demonstrates the robustness of CATNet through controlled noise injection experiments. By introducing heading perturbations (reflecting model parameter variations) and location offsets (emulating transmission interference), we simulate real-world noise conditions.  Taking OPV2V as the representative case, two key observations emerge: 1) Progressive performance degradation in baseline methods under escalating noise intensities, with maximum $AP@0.7$ degradation of 7.98\%/10.02\% in heading and positional noise;  2) CATNet maintains consistent accuracy with only a 0.6\% degradation in AP@0.7, demonstrating effective suppression of types of noise sources: spatial misalignment caused by transmission noise while mitigating vehicle heading deviations induced by model parameter sensitivity.

\noindent \textbf{Latency Robustness Evaluation.} We assess latency robustness by subjecting methods to progressively harsher asynchronous conditions on OPV2V. For each condition, collaborators are perturbed with pose noise ($\sigma=0.2/0.2$) and a random latency drawn from a uniform distribution $U[0, L]$, where we systematically increase the maximum latency $L$ from 200ms to 500ms. The results in Table~\ref{tab:latency_evaluation} show that while performance degrades universally with increasing $L$, CATNet maintains a significant performance advantage across all conditions. This demonstrates the effectiveness of our temporal synchronization mechanism and validates the robustness against unpredictable delays.

% \begin{table}[t] 
% \centering
% \begin{tabular}{@{} l | c c @{}} % 移除了 V2XSet 的列
% \toprule
% Configuration & OPV2V & DAIR-V2X \\ % 移除了 V2XSet 的表头
% \midrule
% Baseline & 0.738\,/\,0.509 & 0.625\,/\,0.446 \\ 
% + STSync & 0.818\,/\,0.678 & 0.683\,/\,0.496 \\ 
% + STSync + WTDen & 0.834\,/\,0.680 & 0.717\,/\,0.549 \\ 
% CATNet & \textbf{0.843\,/\,0.686} & \textbf{0.723\,/\,0.565} \\ 
% \bottomrule
% \end{tabular}
% \caption{Performance comparison of different module combinations on the OPV2V and DAIR-V2X datasets. This ablation study demonstrates the effectiveness of each proposed component.} % 更新了标题，移除了 V2XSet
% \label{tab:effectiveness}
% \end{table}

\begin{table}[t] 
\caption{Performance comparison of different module combinations on the OPV2V and DAIR-V2X datasets. This ablation study demonstrates the effectiveness of each proposed component.} % 更新了标题，移除了 V2XSet
\centering
\begin{tabular}{@{} l | c c @{}} % 移除了 V2XSet 的列
\hline
Configuration & OPV2V & DAIR-V2X \\ % 移除了 V2XSet 的表头
\hline
%Baseline & 0.738\,/\,0.509 & 0.625\,/\,0.446 \\ 
 Baseline & 0.595\,/\,0.384 & 0.659\,/\,0.461 \\ % 新增行
 + STSync & 0.818\,/\,0.678 & 0.683\,/\,0.496 \\ 
 + WTDen  & 0.645\,/\,0.461 & 0.671\,/\,0.482 \\
 + AdpSel  & 0.624\,/\,0.425 & 0.666\,/\,0.477 \\
 + STSync + WTDen & 0.834\,/\,0.680 & 0.717\,/\,0.549 \\ 

 + STSync + AdpSel & 0.822\,/\,0.682 & 0.708\,/\,0.553 \\ % 新增行
CATNet & \textbf{0.843\,/\,0.686} & \textbf{0.723\,/\,0.565} \\ 
\hline
\end{tabular}
\label{tab:effectiveness}
\end{table}

\subsection{Ablation Studies and Analysis}

\noindent \textbf{Effectiveness of Core Components.} We conducted a comprehensive ablation study to evaluate the contribution of each core component across three datasets. We select the standard intermediate method \cite{xu2022opv2v} as our baseline. As presented in Table \ref{tab:effectiveness}, all three modules consistently enhanced detection accuracy. Notably, STSync delivered the most significant improvements. For instance, it achieved gains of 22.3\% and 2.4\% in AP@0.5 compared to the baseline on the OPV2V and DAIR-V2X datasets, respectively. The final integrated CATNet model exhibited overall accuracy gains of 24.8\% and 6.4\% in AP@0.5 over the baseline.

\begin{figure}[!t]
    \centering % 使图片居中显示
    \includegraphics[width=0.48\textwidth]{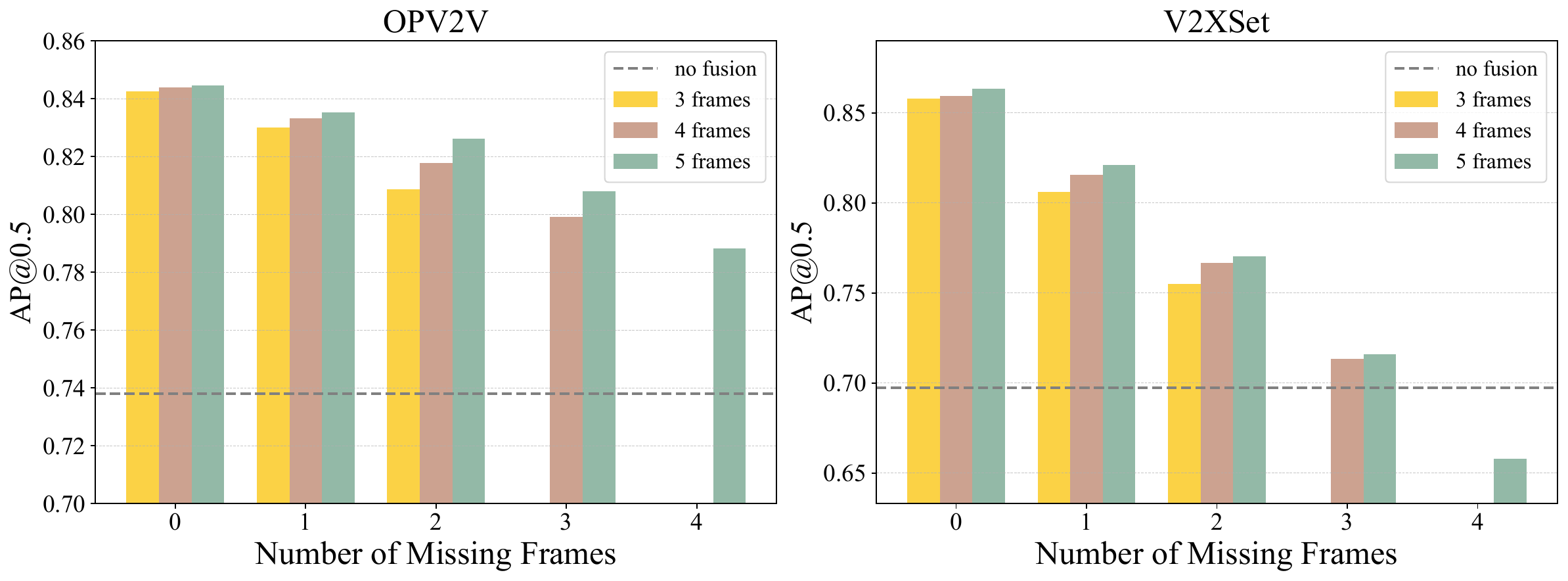}
    \caption{Comparison of STSync under historical data loss.} % 为图片添加标题
    \label{fig:hisloss} % 为图片添加标签，方便引用
\end{figure}

% 突发性通信中断！
\noindent \textbf{Robustness to Incomplete Historical Data.} To evaluate robustness against information loss, we simulated communication outages by randomly dropping a portion of data packets received from collaborators within a historical time window (e.g., 600ms). Results in Figure \ref{fig:hisloss} reveal that the model maintained strong performance even under extreme conditions. On OPV2V, the performance consistently remained above 78\%, and on V2XSet, the model still achieved over 65.0\% AP@0.5 even under delays of up to 600ms. These results demonstrate that CATNet exhibits excellent robustness in edge scenarios with high latency.

\noindent \textbf{Token Retention Analysis.} Table \ref{tab:proportion} analyzes the effect of the token retention ratio ($k\%$) in the standalone AdpSel module under ideal conditions. Results show that a 0.3 retention ratio achieves optimal performance on the dataset, emphasizing the critical role of optimal parameter selection in enhancing model accuracy.

% proportion
\begin{table}[!tb]
\caption{Performance comparison of AdpSel with varying configuration ratios on the OPV2V, V2XSet, and DAIR-V2X datasets.}
  \centering
    \begin{tabular}{cccc}
    \hline
    {Proportion} & OPV2V & V2XSet & DAIR-V2X \\
    \hline
    0.1   & 0.854/0.674 & 0.873/0.717 & 0.767/0.589 \\
    0.2   & 0.854/0.674 & 0.873/0.716 & 0.774/0.587 \\
    \textbf{0.3}   & \textbf{0.855}/\textbf{0.691} &\textbf{0.874}/\textbf{0.717} & \textbf{0.775}/0.603 \\
    0.4   & 0.854/0.675 & 0.873/0.716 & 0.773/0.604 \\
    0.5   & 0.845/0.681 & 0.863/0.711 & 0.770/\textbf{0.605} \\
    0.6   & 0.836/0.663 & 0.863/0.702 & 0.765/0.596 \\
    \hline
    \end{tabular}%
  \label{tab:proportion}
\end{table}%

\begin{table}[t] % 将 [h!] 更改为 [t]，这是AAAI更常用和推荐的浮动位置 。AAAI对 [h!] 等强制位置参数没有明确提及，但 [t] 更符合其建议。
\setlength{\tabcolsep}{6pt} % 调整列间距。如果表格太宽，可以减小此值，AAAI建议1mm 。
\caption{AdpSel performance with block masking on OPV2V.}
\centering
\begin{tabular}{@{}l|ccc@{}} % 移除了所有竖线，因为AAAI的示例和建议中通常不使用表格内部竖线。
\hline % 替换 \hline 为 \toprule 
Module / Metric & AP@0.3 & AP@0.5 & AP@0.7 \\
\hline % 替换 \hline 为 \midrule 
AdpSel & 0.905 & 0.897 & 0.791 \\
AdpSel (high mask) & 0.366 & 0.364 & 0.328 \\
AdpSel (low mask) & 0.791 & 0.784 & 0.707 \\
\hline % 替换 \hline 为 \bottomrule 
\end{tabular}
\label{tab:adpsel_masking}
\end{table}

\noindent \textbf{Effectiveness of AdpSel.} 
% We analyzed impact of AdpSel within the OPV2V dataset by applying equal-intensity noise masks to high- and low-attention regions. As shown in Table \ref{tab:adpsel_masking}, masking high-attention areas significantly drops detection accuracy, underscoring their critical role. The fusion of Primary and Secondary features achieves the highest feature reinforcement, vital for comprehensive environmental understanding and precise target localization in cooperative perception. These findings highlight AdpSel's importance in guiding attention to salient features, enhancing situational awareness and detection accuracy.
Table \ref{tab:adpsel_masking} demonstrates AdpSel's impact by applying equal-intensity noise masks to high- and low-attention regions. The results show that masking high-attention areas significantly drops detection accuracy, confirming their critical role. Furthermore, the fusion of Primary and Secondary features achieves the highest feature reinforcement, which is vital for comprehensive perception. These findings highlight AdpSel's effectiveness in guiding attention to salient features.

\begin{table}[t] 
\caption{Performance of Different Methods under Mixed Localization and Heading Noise} 
    \centering
    \label{tab:performance_noise}
    \begin{tabular}{lccc}
        \hline
        Method & 0/0 & 0.2/0.2 & 0.4/0.4 \\
        \hline
        Agent-graph & 0.872/0.790 & 0.868/0.737 & 0.838/0.588 \\
        WTDen     & 0.893/0.795 & 0.875/0.737 & 0.860/0.654 \\
        \hline
        FeaCo           & 0.881/0.784 & 0.874/0.755 & 0.855/0.649 \\
        CoAlign         & 0.888/0.790 & 0.881/0.748 & 0.848/0.633 \\
        DSRC            & 0.894/0.803 & 0.887/0.752 & 0.854/0.638 \\
        CATNet          & 0.905/0.807 & 0.896/0.763 & 0.870/0.663 \\
        \hline
    \end{tabular}
    \label{tab:WTDen_noise}
\end{table}

\noindent \textbf{Effectiveness of WTDen.} The WTDen module suppresses global noise from STSync by processing features in the wavelet domain. It simultaneously corrects feature misalignment via two branches: Wavelet Mamba uses our innovative hybrid scan to capture global positional relationships from frequency perspectives for alignment, while Wavelet Conv models local features to reduce inconsistencies. Table \ref{tab:WTDen_noise} demonstrates this module's effectiveness. In delay-free scenarios with mixed localization and heading noise, CATNet surpasses existing noise-robust SOTA methods, especially under extreme conditions, confirming its crucial robustness for collaborative perception.

\begin{figure}[!t]
    \centering % 使图片居中显示
   \includegraphics[width=0.48\textwidth]{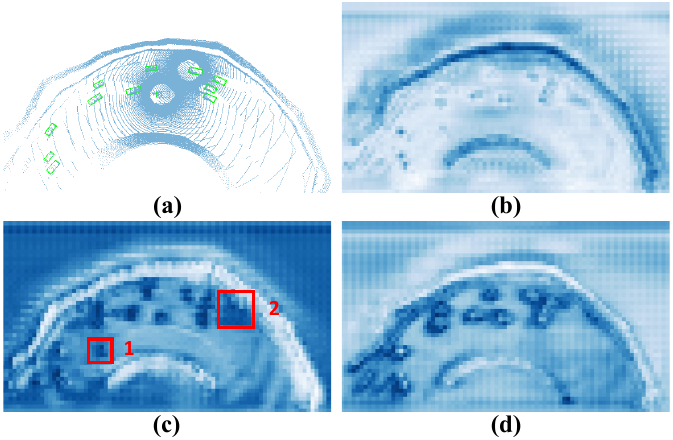} % 插入图片，调整宽度为当前栏宽度的 0.9 倍
    \caption{Heatmap comparison of collaborative perception performance. (a) Ground-truth. (b) Baseline. (c) CATNet (w/o WTDen). (d) CATNet (w/ WTDen).} % 为图片添加标题
    \label{fig:Heatmap} % 为图片添加标签，方便引用
\end{figure}

\subsection{Qualitative Results}

Figure \ref{fig:Heatmap} visualizes the impact of WTDen through four cases. The proposed module significantly reduces artifacts in Regions 1-2 by synergistically
mitigating both sensor noise and network-induced artifacts. The clearer denoised features in Figure \ref{fig:Heatmap} (d) further validate its ability to suppress interference while preserving structural semantics.

\section{Conclusion}

We investigate the issue of latency and noise in cooperative perception systems and propose a robust collaborative network, CATNet. This network first aligns asynchronous multi-vehicle features via a temporal recurrent module, then performs a comprehensive purification by first cleaning signal-level distortions and subsequently using semantic coherence to guide the final adaptive fusion. Results show that CATNet performs well in complex scenarios.

\noindent \textbf{Acknowledgement.} The work is supported in part by the S\&T Program of Hebei Province (Beijing-Tianjin-Hebei Collaborative Innovation Special Program) under Grant 25240701D.

{
    \small
    \bibliographystyle{ieeenat_fullname}
    \bibliography{main}
}

\end{document}